\documentclass[conference]{IEEEtran}
\IEEEoverridecommandlockouts
\usepackage{cite}
\usepackage{amsmath,amssymb,amsfonts}
\usepackage{algorithmic}
\usepackage{graphicx}
\usepackage{textcomp}
\usepackage{xcolor}
\usepackage{latexsym}
\usepackage{amssymb}
\usepackage{amsmath}
\usepackage{amsthm}
\usepackage{booktabs}
\usepackage{enumitem}
\usepackage{graphicx}
\usepackage{color}
\usepackage{multirow}
\usepackage{textcmds}
\usepackage{url}
\usepackage{microtype}

\def\BibTeX{{\rm B\kern-.05em{\sc i\kern-.025em b}\kern-.08em
    T\kern-.1667em\lower.7ex\hbox{E}\kern-.125emX}}
\begin{document}

\title{Trustworthy Compression? Impact of AI-based Codecs on Biometrics for Law Enforcement \\
}

\author{Sandra Bergmann$^1$, Denise Moussa$^1$ and Christian Riess$^1$ \\
	$^1$Friedrich-Alexander University Erlangen-Nürnberg, Germany\\
	\{sandra.daniela.bergmann, denise.moussa, christian.riess\}@fau.de
}

\maketitle

\begin{abstract}

	Image-based biometrics can aid law enforcement in various aspects, for example
	in iris, fingerprint and soft-biometric recognition.
	A critical precondition for recognition is the availability of sufficient
	biometric information in images.
	It is visually apparent that strong JPEG compression removes such details.
	However, latest AI-based image compression seemingly preserves many image details even for very strong compression factors.
	Yet, these perceived details are not necessarily grounded in measurements,
	which raises the question whether these images can still be used for biometric
	recognition. 
	
	In this work, we investigate how AI compression impacts iris, fingerprint and soft-biometric (fabrics and tattoo) images.
	We also investigate the recognition performance for iris and fingerprint images after AI compression.
	It turns out that iris recognition can be strongly affected, while 
	fingerprint recognition is quite robust.
	The loss of detail is qualitatively best seen in fabrics and tattoos images.
	Overall, our results show that AI-compression still permits many biometric
	tasks, but attention to strong compression factors in sensitive tasks is
	advisable.
\end{abstract}


\section{Introduction}
Biometrics can serve various purposes in law enforcement and the wider security sphere, e.g., to add or remove persons from a list of possible suspects, to control access, or to gain investigative cues. Important primary biometric data to identify individuals are fingerprints or irises~\cite{Jain07}. Fingerprints are frequently used either to identify repeat offenders~\cite{Jain10, Jain11}.  
Iris images may also provide other helpful information about a person like gender, age, eye color or ethnicity~\cite{Dantcheva16}. 
Furthermore, so-called ``soft-biometrics'', like clothing (texture or color) or tattoos (position or content), are also essential cues to characterize individuals~\cite{Lee08, Nixon15}. Such cues are particularly useful when primary biometric data is not available~\cite{Dantcheva16}.

Biometric recognition greatly benefits from high-quality data.  However, such
high-resolution images are not always available considering, e.g., images from
surveillance cameras.
It is well-known that strong (lossy) compression, like JPEG, can negatively affect
the biometric recognition of individuals~\cite{Ives05,Daugman08,Mascher-Kampfer07}.
For instance, strong compression increases the false negative matches for iris recognition~\cite{Matschitsch07}, which questions the usefulness of such images for law enforcement.

\begin{figure}[!h]
	\centerline{\includegraphics[keepaspectratio,width=\linewidth,height=15cm]{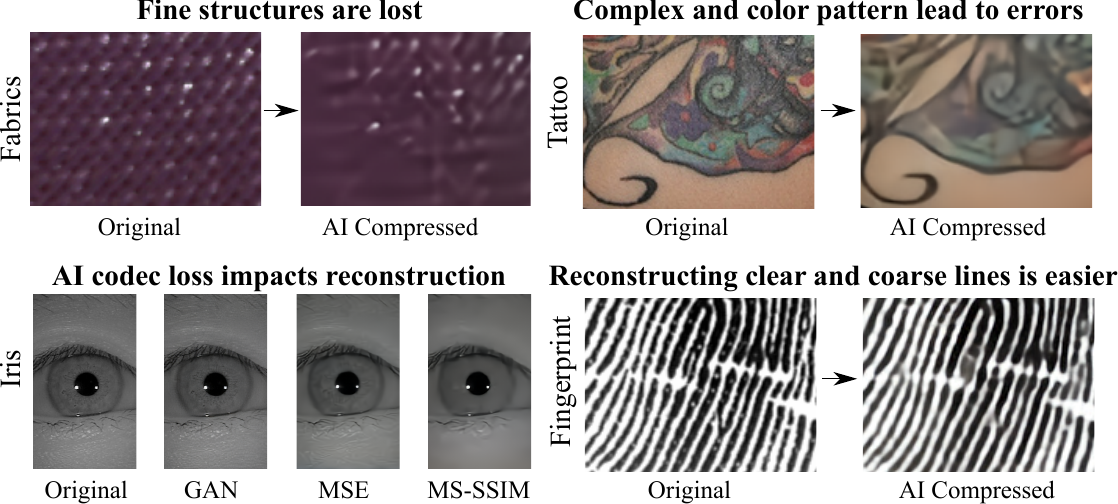}}
	\caption{AI codecs can impact typical biometric data for law enforcement depending on the use case. Best viewed in its digital version.}
	\label{fig_cover_image}
\end{figure}

AI-based compression recently emerged as a new way to store images with
considerably higher quality at lower bitrates than
JPEG~\cite{Toderici17,Balle18,Mentzer20}. The JPEG committee is currently standardizing the JPEG AI format~\cite{JPEGAI_Ascenso}.
This raises the question how
typical biometric tasks perform on such AI compressed images. On one hand, one may expect that the bitrate gain aids biometric recognition. On the other hand, AI compression is based on generative AI, which may introduce new challenges.
For example, it is possible that compression alters specific biometric traits,
which was recently reported for certain combinations of segmentation and
compression algorithms in iris recognition for identification
systems~\cite{Jalilian22}.

In this paper, we provide an overview of the impact of AI
compression for typical biometric tasks in law enforcement, namely on
images of the irises, fingerprints and soft-biometrics (fabrics and tattoos).
For all modalities, we analyze the impact of AI compression on image quality. For
irises and fingerprints, we also evaluate the impact on biometric recognition.
Our work shows that important biometric characteristics may be lost or altered
for AI-compressed images, depending on the biometric task.
Yet, this may be not obvious to the eye, since even strong AI compression
oftentimes appears to be of superior perceptual quality.
Selected  findings of this work are:
\begin{enumerate}
	\item The structural similarity (SSIM) between uncompressed and AI-compressed biometric images is lower, especially for middle and high quality levels, than for JPEG-compressed images.
	\item The impact of AI-based compression on biometric images depends on the special characteristics of its data. Figure~\ref{fig_cover_image} shows that fine structures are lost (iris and fabrics); while the reconstruction of complex patterns and color transitions is not fully successful (tattoo), the reconstruction of clear and coarser lines is more successful (fingerprint).
	\item The loss function of an AI compressor impacts the reconstruction of the biometric traits (see Fig.~\ref{fig_cover_image}). MSE loss has a greater influence on iris and fabrics images, while MS-SSIM loss has greater influence on fingerprint and tattoo images. GAN-loss provides high perceptual quality, but worse biometric features at high quality levels.
\end{enumerate}

The paper is organized as follows. Section 2 and 3 outline the related work and foundations of AI compression, respectively. Section 4 studies the impact of AI compression on iris, fingerprint, and soft-biometrics. Section 5 concludes this work.


\section{Related Work}
Biometrics research is extensive, which inevitably limits the scope of this review.  
The foundations of iris recognition were laid by Daugman~\cite{Daugman93,Daugman04} and are still widely employed today. Those works showed that so-called ``iris codes'', a two-bit representation of the iris, are useful to compare two irises. Over the past few decades, deep learning has been extensively used in iris recognition~\cite{Nguyen24, Gao24}, e.g., for iris segmentation~\cite{Kuehlkamp,Lazarski22} or feature extraction~\cite{Minaee16,Yang21}.

Many state-of-the-art fingerprint recognition approaches are based on the extraction of minutiae. Traditional methods extract minutiae with, e.g., skeleton extraction~\cite{Farina99} or direct detection approaches~\cite{Maio97}. More recently, minutiae were extracted with deep learning~\cite{Tang17,Nguyen19}.

Soft-biometrics, like clothings or tattoos, are important to identify individuals~\cite{Nixon15}. Earlier works on tattoo recognition use SIFT features to localize characteristics of a tattoo and leveraged a matching algorithm to measure the similarity of two tattoos~\cite{Lee12,Han13}. Later, only few deep learning approaches emerged~\cite{Di17,NicolasDiaz22} due to a lack of suitable datasets~\cite{gonzalezsoler2024}.

Due to the importance of unique image details for biometric recognition, much research addresses image quality, particularly when using standard JPEG compression~\cite{Daugman08}.
For iris recognition, numerous studies demonstrate that strong JPEG compression results in fewer true positives, while the number of false positives remains unchanged~\cite{Ives05,Matschitsch07}.
Mascher-Kampfer \textit{et al.} show that JPEG compression can dramatically decrease fingerprint similarity, which also weakens fingerprint recognition~\cite{Mascher-Kampfer07}.

Although research on the effect of AI compression on image forensics is still scarce when compared to JPEG, some work exists already. Those approaches show that AI compression challenge image forensics tools and watermarking~\cite{Bhowmik21,Berthet22,Berthet23}, while
Bergmann~\textit{et al.} reveal traces of AI compression in the frequency and spatial domain~\cite{Bergmann23,Bergmann24}.

In biometrics, AI compression on iris identification systems has been studied by Jalilian \textit{et al.}~\cite{Jalilian22}. They compare different AI codecs and show that AI compression can affect iris recognition depending on the segmentation algorithm and compression strength.
In contrast to that, our work provides an overview of the impact of AI compression on biometric data for law enforcement considering iris, fingerprint and soft-biometrics.
We analyze AI-compressed images with respect to their visual image quality, their structural similarity when compared to their originals, and their impact on recognition. 

\section{Neural Network Compression}

Toderici \textit{et al.} introduced AI compression based on recurrent neural networks~\cite{Toderici17}. Further works use autoencoders to encode and reconstruct images~\cite{Theis17, Balle18}.
Ball\'{e}~\textit{et al.} use a hyperprior to model the probability density of the latents~\cite{Balle18}.
The performance of hyperprior models can be further improved by combining them with spatially autoregressive models~\cite{Minnen18} or with channel-conditional models~\cite{Minnen20}.
Furthermore, Mentzer \textit{et al.} present impressive results with a GAN~\cite{Mentzer20} and Hoogeboom \textit{et al.}
show the use of diffusion models for AI compression~\cite{Hoogeboom23}.

The architecture of AI compression algorithms shares some similarities.
A convolutional encoder transforms an image $\mathbf{x}$ into the quantized latent space $\mathbf{y} = E(\mathbf{x})$.
The latent space represents the spatial dependencies in an image, and hyperpriors $\mathbf{z}$ guide this representation~\cite{Balle18,Duan22}. 
To decompress the image, a convolutional decoder uses $\mathbf{z}$ to reconstruct the image.

All components of the codecs are trained together to achieve the best rate-distortion loss function.
The choice of the neural network loss encourages specific properties of the compression algorithm.
For example, the mean square error (MSE) loss is popular, but prone to suppress visual details. Other loss functions encourage the preservation of details in texture and contrast, such as the MS-SSIM
loss~\cite{Mohammadi23}.
Furthermore, GAN-based image compressors learn a loss function, which enables the
reconstruction of images of high visual quality~\cite{Mentzer20}.
However, these compressors may exhibit color shifts or the replacement of textures during the reconstruction. Hence, each loss leads to a different variation between the original (uncompressed) image and the decompressed image~\cite{Mohammadi23}.

\section{AI Compression in Biometrics}
Our study uses three AI compression methods in five variants.
The methods are ``Hific'' as a GAN-based codec~\cite{Mentzer20}, ``Mbt'' as an autoregressive hyperprior model~\cite{Minnen18}, and ``Ms'' as a channel-wise autoregressive model~\cite{Minnen20}.
For Mbt and Ms, we also differentiate between the trained models with MSE loss (Mbt-mse, Ms-mse), and variants that are trained with MS-SSIM loss (Mbt-msssim, Ms-msssim). For compression, we use the ``Tensorflow Compression'' framework ~\cite{tfc_github}.
Each model accepts a quality level parameter.
Hific provides quality levels of \{lo, mi, hi\}, where ``hi'' indicates the highest image quality, i.e., the most faithful representation of the image.
Mbt-mse and Mbt-msssim use the quality levels \{1-8\}, 
Ms-mse \{1-10\} and Ms-msssim 
\{1-9\}, where higher values indicate higher image quality.
To meet the required input format for AI compression, grayscale images are expanded to grayscale RGB.
The bitrate of AI-compressed images is about an order of magnitude lower than for JPEG images. 
Nevertheless, as a baseline, we also use JPEG images with quality levels \{5, 10, 25, 50, 75, 90, 95\}.

We investigate the impact of AI compression on irises, fingerprints and soft-biometrics images. Due to space constraints, and the similarity of results for Mbt and Ms, we only visualize Mbt, but provide all results on our website.

\subsection{Iris Recognition}
\label{sec_irisrecognition}
We first introduce the iris recognition algorithm, and then quantitatively and qualitatively analyze the quality of AI-compressed images.
Finally, we evaluate the influence of AI compression on the selected iris recognition algorithm. 

\subsubsection{Algorithm}
\label{sec_worldcoin}
The experiment uses the open-source iris recognition framework from Worldcoin~\cite{Worldcoin}, which generates iris codes according to Daugman~\cite{Daugman93}.
The framework consists of four main steps. First, the iris is segmented with a deep neural network by Lazarski~\textit{et al.}~\cite{Lazarski22}.  
Second, the iris region is normalized to extract the most relevant iris textures and transforms them from cartesian to polar coordinates. Third, the iris code is extracted from the normalized iris region with a series of Gabor filters~\cite{Daugman93,Daugman04}. 
Fourth, two iris codes are compared by their Hamming distance, and a match is found if that distance does not exceed $0.38$.

\subsubsection{Evaluation}
The experiment is performed on the CASIA Iris-Thousand-V4 dataset~\cite{CASIAIris}, which consists of grayscale JPEG images of the left and right eyes. We use 5 images each of both eyes for 10 individuals. 
Each image is re-compressed with the AI codecs and with the JPEG algorithm at various quality levels as stated before. For iris recognition we have to check a total of 4950 combinations for each codec.  

\begin{figure}[tb]
	\centerline{\includegraphics[keepaspectratio,width=\linewidth,height=\linewidth]{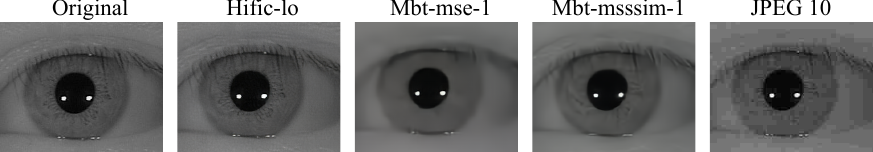}}
	\caption{An iris example from the CASIA-Iris-Thousand dataset~\cite{CASIAIris} compressed at the lowest quality level using different AI Codecs and JPEG.}
	\label{fig_iris_examples}
\end{figure}

\paragraph{Visual Analysis} 
AI-compressed images exhibit a notably higher visual quality than their JPEG-compressed counterparts. Figure~\ref{fig_iris_examples} shows an iris using different codecs at their lowest quality level.
The image quality of AI-compressed images is high, despite the loss of features in some cases within the iris. The images of Hific-lo and Mbt-msssim-1 exhibit details in the iris that look similar to the original. The AI codecs with MSE loss are blurry and leave almost no biometric features. Still, the superior image quality appears as better suited than the JPEG image with block artifacts.

To confirm our visual results, we assess how similar the original images are to their compressed versions and calculate the average SSIM.
Table~\ref{tab_SSIM}~A) shows that for each compression method, decreasing the quality level also decreases the SSIM.
Perhaps surprisingly, at its highest quality level, the SSIM of AI compression is lower than for JPEG.
This indicates that although AI-compressed images exhibit high visual quality, they discard or alter some actual (real) details of the original image.
The GAN-based Hific images show high perceptual quality and have lower SSIM values than the other AI-compression methods. As expected, models trained with MS-SSIM loss perform better than models trained with MSE loss.

\begin{figure}[tb]
	\centerline{\includegraphics[keepaspectratio,width=8.5cm,height=5cm]{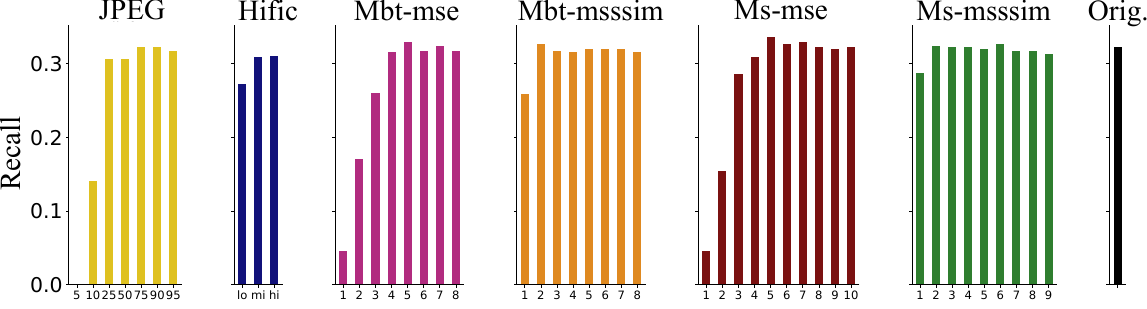}}
	\caption{Recall for iris recognition on AI-based codecs and JPEG.}
	\label{fig_iris_prec_reca}
\end{figure}

\paragraph{Recognition Task} 
\label{sec_classification_iris}
The recognition experiment also uses the CASIA Iris-Thousand-V4 dataset. 
The performance is reported as precision and recall to alleviate the imbalance in the number of non-matching combination to matching combinations.
We use the open-source iris recognition software from Worldcoin for the actual recognition (see Sec.~\ref{sec_worldcoin}). 

This experiment achieves a precision of 1 almost everywhere, i.e., there are no false positives except at JPEG quality level 5, where the algorithm completely fails.
The recall is more interesting, and therefore shown in Fig.~\ref{fig_iris_prec_reca}. The AI compression impacts the number of false negatives. The recall of all methods decreases with increasing compression strength, which is expected.
Particularly noteworthy are AI codecs with MSE loss (Mbt-mse and Ms-mse), where
low recall values occur already at higher quality than for the other methods.
The AI compression level 1 produces more false negatives than JPEG with quality
level 10, which analogously demonstrates that perceived visual
quality does not necessarily correspond to an accurate reconstruction of the
biometric features.  

\begin{table}[!t]
	\renewcommand{\arraystretch}{0.75}
	\caption{Avg. SSIM between original images and compressed versions for iris (a), fingerprint (b), fabrics (c) and tattoo images (d).}
	\label{tab_SSIM}
	\centering
	\begin{tabular}{lccccccccc}
		\toprule
		& A) \textbf{Iris} & B) \textbf{Fprint} & C) \textbf{Fabrics} & D) \textbf{Tattoo}\\
		\midrule
		\multicolumn{4}{l}{\textbf{JPEG}} \\
		Quality level: 5 & 0.787 & 0.929 & 0.713 & 0.687\\
		Quality level:  50 & 0.945 & 0.989 & 0.960 & 0.904\\
		Quality level:  95 & 0.999 & 0.999 & 0.993 &  0.996\\
		\midrule
		\multicolumn{4}{l}{\textbf{Hific}} \\
		Quality level: lo & 0.852 & 0.967 & 0.889 & 0.778 \\
		Quality level: mi & 0.906 & 0.984 & 0.939 & 0.833\\
		Quality level: hi & 0.925 & 0.990 & 0.956 & 0.871 \\
		\midrule
		\multicolumn{4}{l}{\textbf{Mbt-mse}} \\
		Quality level: 1 & 0.867 & 0.938 & 0.823 & 0.774\\
		Quality level: 4 & 0.905 & 0.985 & 0.938 & 0.874\\
		Quality level: 8 & 0.972 & 0.993 & 0.987 & 0.969 \\
		\midrule
		\multicolumn{4}{l}{\textbf{Mbt-msssim}} \\
		Quality level: 1 & 0.889 & 0.906 & 0.856 & 0.769\\
		Quality level: 4 & 0.944 & 0.966 & 0.956 & 0.876\\
		Quality level: 8 & 0.990 & 0.984 & 0.990 & 0.971\\
		\midrule
		\multicolumn{4}{l}{\textbf{Ms-mse}} \\
		Quality level: 1 & 0.871 & 0.964 & 0.837 & 0.803\\
		Quality level: 5 & 0.927 & 0.990 & 0.963 & 0.917\\
		Quality level: 10 & 0.991 & 0.994 & 0.995 & 0.986\\
		\midrule
		\multicolumn{4}{l}{\textbf{Ms-msssim}} \\
		Quality level: 1 & 0.898 & 0.920 & 0.875 & 0.793 \\
		Quality level: 5 & 0.964 & 0.978 & 0.971 & 0.904\\
		Quality level: 9 & 0.994 & 0.983 & 0.994 & 0.963\\
		\bottomrule
	\end{tabular}
\end{table}

\subsubsection{Impact on Iris Codes Distance}
The increased number of false negatives for lower-quality AI compression can be attributed to an increase of the distances between iris codes. 
It is instructive to visualize the change of distances between matches and non-matches to better understand its impact.

\begin{figure}[tb]
	\centerline{\includegraphics[keepaspectratio,width=8.5cm,height=8.5cm]{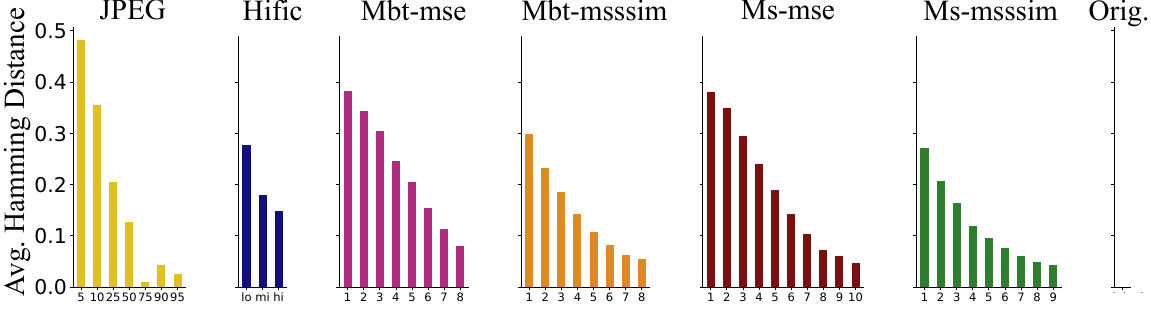}}
	\caption{Average Hamming distance from two iris codes of the same original and compressed image.}
	\label{fig_iris_distance_same}
\end{figure}

To this end, we calculate the Hamming distances between pairs of original images and their compressed counter-parts. Figure~\ref{fig_iris_distance_same} shows the averages of these distances per compression method. Overall, the distances increase with stronger compression for all compression methods, and they shift closer to the threshold of $0.38$. Not surprisingly, JPEG quality level 5 exhibits the largest average Hamming distance.
However, it is surprising that higher-quality AI codecs exhibit larger distances than JPEG. Again, AI codecs with the MSE loss show the largest distances and the GAN-based Hific shows for the highest quality a larger distance than the other codecs. 

\begin{figure}[tb]
	\centerline{\includegraphics[keepaspectratio,width=8.5cm,height=8.5cm]{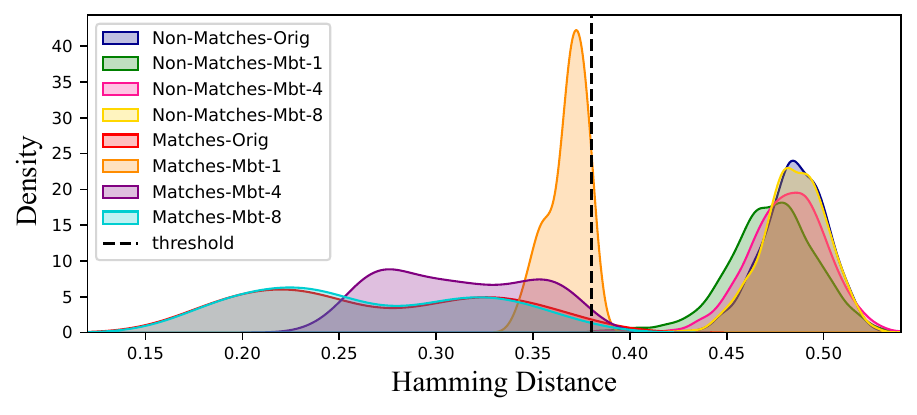}}
	\caption{Distribution of non-matches and matches depending on the distance from iris recognition for an AI codec with different quality levels.}
	\label{fig_iris_distribution_distance}
\end{figure}

The distributions of Hamming distances for non-matches and matches provides
further insights into this behavior.
Since the AI codecs with MSE loss exhibit the largest changes, we present the distributions of non-matches and matches from images from Mbt-mse.
Figure~\ref{fig_iris_distribution_distance} illustrates the distributions of distances between the original and Mbt-mse, with quality levels 1, 4 and 8, respectively. Stronger AI compression shifts the distances of matches closer to the threshold. The distribution of non-matches also shifts slightly towards the matches with stronger compression. Consequently, one can expect that the choice of a suitable threshold for recognition is more challenging for AI-compressed images.
The result will either show more incorrect negative or positive classifications.

Overall, our results show that AI compression methods create images of high visual quality suggesting particular usefulness for law enforcement. However, this appearance is misleading, as AI-compressed images may lack biometric details which are significant for iris recognition.

\begin{figure}[t]
	\centerline{\includegraphics[keepaspectratio,width=\linewidth,height=\linewidth]{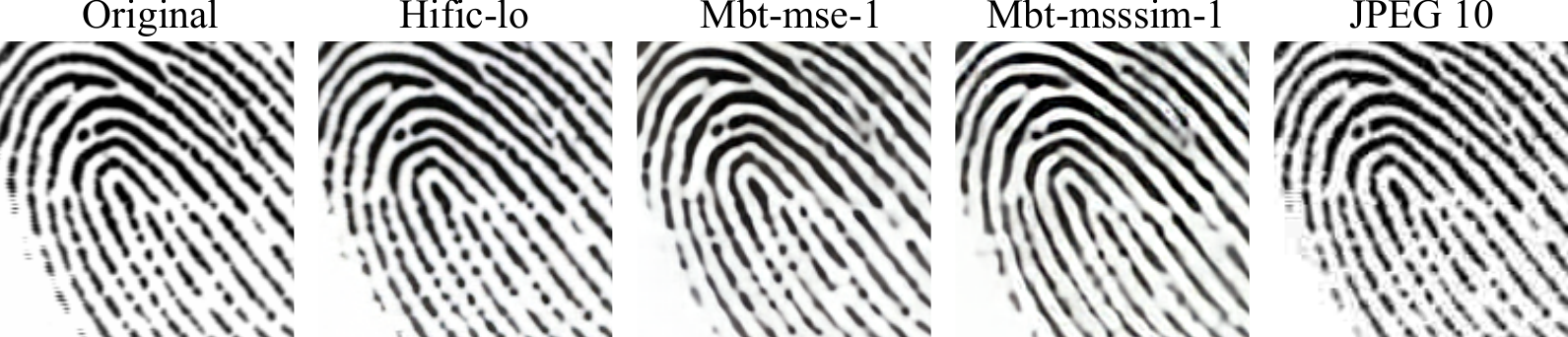}}
	\caption{A fingerprint example from the DB1 FVC2000 dataset~\cite{FVC2000} compressed at the lowest quality level using different AI codecs and JPEG.}
	\label{fig_fprint_examples}
\end{figure}

\subsection{Fingerprint Recognition}
\label{sec_fingerprint}
We first introduce the fingerprint recognition algorithm for the experiments. The analysis consists then of a visual analysis of the compressed images and a study of the structural similarity (SSIM) of the images and their compressed counter-parts. Then, the impact of AI compression on fingerprint recognition performance is quantitatively evaluated. 

\subsubsection{Algorithm}
\label{sec_sourceAFIS}
We use the open-source framework SourceAFIS~\cite{SourceAFIS}.
The algorithm uses fingerprint minutiae, following a traditional image processing pipeline. The image is first filtered, then fingerprint minutiae are detected. Lines between two minutiae are detected, which are referred to as ``edges''. A nearest neighbour search on such an edge acts as a coarse retrieval for a matching fingerprint. A refined comparison checks whether other edges or minutiae also match. The number and the quality of the matches determine the final score. For our investigations, we use a matching threshold of 40 as proposed by the authors.

\subsubsection{Evaluation}
The experiment is performed on the DB1 dataset of the FVC2000 Fingerprint~\cite{FVC2000}. It consists of fingerprint images of 10 people, each with 8 different prints.
The grayscale images are in TIF format. Before compression with the various AI codecs and JPEG, we convert them to three channel PNGs. Once more, all selected quality levels are considered. For fingerprint recognition we have to check a total of 3160 combinations for each codec.

\paragraph{Visual Analysis} 
To highlight the strongest changes from compression, Figure~\ref{fig_fprint_examples} visualizes fingerprints compressed with the lowest quality levels of the codecs.
The AI codecs provide higher image quality than JPEG, but papillary lines of the fingerprints appear slightly altered, yet very close to the original image. The images of Mbt have an additional blurriness. Images with MS-SSIM loss appear more blurred and altered than those with MSE, 
which is in contrast to the observations for iris images.
However, overall, the AI codecs better reproduce the straighter and thicker fingerprint features than fine iris features.  

To follow up on some missing details of the papillary lines, we quantitatively measure the similarity between original and compressed fingerprint images.
Table~\ref{tab_SSIM} B) shows the results of the average SSIM between original and compressed fingerprints.
Again, the similarity of the images decreases with stronger compression and JPEG with quality 95 shows the highest similarity.
In general, the SSIM values are higher than for the iris images. On one hand, this is attributed to the better quality of the source images, and on the other hand to the fingerprint pattern which is better preserved by the AI codecs.
The SSIM also confirms that the images with the MS-SSIM loss models are less similar to the original images than the MSE loss and GAN-based models (see Fig.~\ref{fig_fprint_examples}).
That is unexpected, as the MS-SSIM loss specifically aims to preserve structural similarity, and one would assume that it outperformes the other methods in this metric.

\paragraph{Recognition Task}
The quantitative recognition experiment also uses the DB1 dataset of the FVC2000 Fingerprint Challenge~\cite{FVC2000}.  
We follow the same experimental protocol as for iris recognition: we compare recognition performances for original and compressed images with
the SourceAFIS framework and precision and recall as metrics.

\begin{figure}[tb]
	\centerline{\includegraphics[keepaspectratio,width=9cm,height=8.5cm]{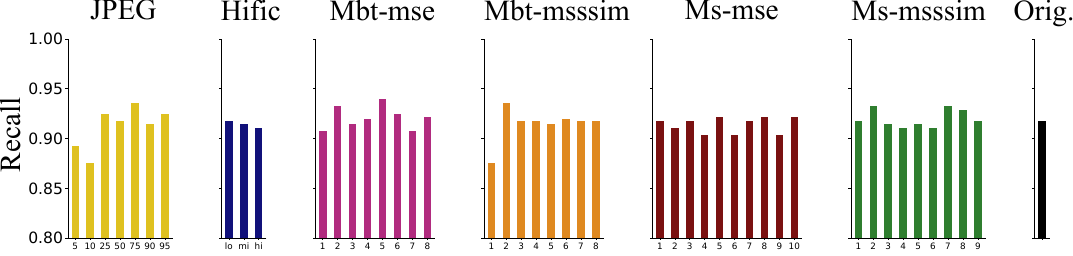}}
	\caption{Recall for fingerprint recognition on AI-based codecs and JPEG.}
	\label{fig_fprint_prec_reca}
\end{figure}

On all image types, a precision of nearly 1 is obtained, analogously to the
iris experiment. Fig.~\ref{fig_fprint_prec_reca} shows the recall. In contrast
to iris recognition, the recall is not much affected by compression.
Only Mbt-msssim-1 and JPEG images at lowest quality levels perform slightly worse. Hence, although the AI compression causes minor alterations of the fingerprint, they do not impair the recognition.  

Overall, our results show that fingerprint recognition is more resilient to AI compression than iris recognition. This is due to AI compression affecting fine iris traits more negatively than the coarser and straighter papillary lines of a fingerprint.

\begin{figure}[tb]
	\centerline{\includegraphics[keepaspectratio,width=\linewidth,height=\linewidth]{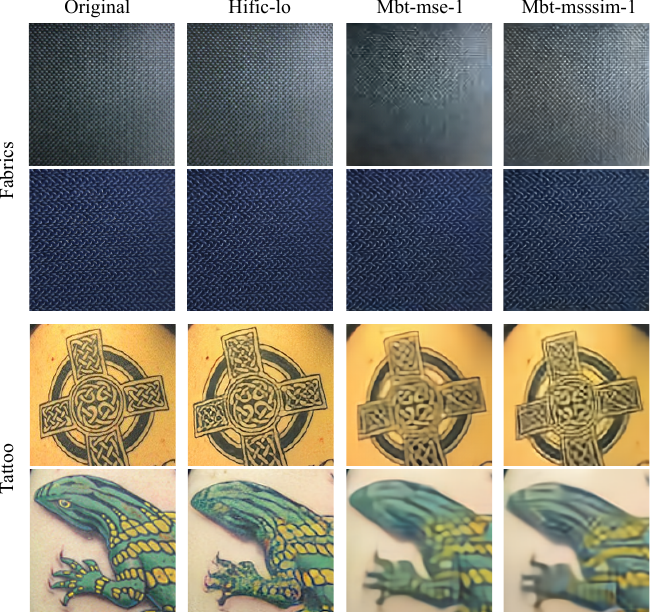}}
	\caption{Fabric (top and second row) and tattoo (third and bottom row) examples from The Fabrics dataset~\cite{FabricsDataset} and DeMSI dataset~\cite{Hrka16}. The images are compressed with different AI codecs on the lowest quality level. This image is best viewed in its digital version, contrast-enhanced for better visualization.}
	\label{fig_fabrics_tattoo_example}
\end{figure}

\subsection{Soft-Biometrics}
Clothing patterns and tattoos are important soft-biometrics. To investigate the impact of AI compression on such images, we use 200 RGB fabric images from The Fabrics Dataset~\cite{FabricsDataset} and 60 RGB images of different tattoos from the DeMSI dataset~\cite{Hrka16}. Again, we compress the images using the selected quality levels and codecs.

Figure~\ref{fig_fabrics_tattoo_example} shows soft-biometrics examples compressed with different AI codecs at the lowest quality level. For fabrics, we observe that fine patterns, like nylon, are more difficult to deal with for AI compression (see Fig.~\ref{fig_fabrics_tattoo_example} top row). Analogously to the findings on iris images, the AI codecs with MSE loss fail on that task. All compressors work much better for coarser patterns as found in the polyester sample, even though some details are lost here as well (see Fig.~\ref{fig_fabrics_tattoo_example} second row). For tattoo images, we observe that more complex patterns, like the cross with entanglements (see Fig.~\ref{fig_fabrics_tattoo_example} third row), are more challenging for AI compression.
The GAN-based Hific shows the best perceptual quality of the entanglements.
Particularly, Mbt has difficulties to reconstruct the entanglements of the cross for both losses.
Nevertheless, the structure with MSE looks slightly more accurate, which is somewhat unexpected considering the nature of the loss function (this observation is analogous to MSE-based compressors on fingerprints in Sec.~\ref{sec_fingerprint}).
The bottom row of Figure~\ref{fig_fabrics_tattoo_example} shows that color transitions can challenge AI codecs as well.
For example, the AI codecs fail to reconstruct the yellow, black and green structures of the lizard (e.g., misses the yellow eye).
However, we only observe a clearly missing yellow eye for the strongest compression when using GAN-based and MSE codecs while for MS-SSIM codecs, the yellow eye is only clearly visible starting from quality level 4.

As previously, we examine the average SSIM between the original soft-biometric images and their compressed versions.
Table~\ref{tab_SSIM} C) contains the results for fabrics and Table~\ref{tab_SSIM} D) those for tattoos.
As expected, the SSIM decreases with stronger compression of fabric and tattoo images.
Furthermore, the medium and highest AI compressed images show again lower SSIM results than JPEG.
For fabrics the SSIM analysis confirms that AI codecs with MSE loss perform worse than MS-SSIM.
Notably, the GAN-based Hific tends to have lower SSIM values at medium and highest quality when compared to the other codecs, a phenomenon that we already observed for iris images in Tab.~\ref{tab_SSIM}~A).
For tattoo images, the SSIM values are lower than for the other images.
Other than that, analogous observations can be made as for fabrics: AI codecs trained with MS-SSIM exhibit slightly lower SSIM than codecs trained with MSE loss, and again
the GAN-based Hific is visually more convincing, but achieves lower SSIM scores for medium and high quality. This fact demonstrates that the perceptual fidelity and realism of AI compression is very high, but the reconstruction can deviate further from the input.

\section{Conclusion}
This work analyzes the impact of AI compression on iris, fingerprint and soft-biometrics (fabric and tattoo) images for law enforcement tasks.
We investigate the influence of AI compression quantitatively and qualitatively by assessing the structural and perceptual similarity of compressed and uncompressed images.
For iris and fingerprint images, we additionally evaluate the impact on the recognition performance.

Our results show that AI-compressed images
\textit{appear} to contain a large amount of details, but these details do not
necessarily carry biometric information. Somewhat surprisingly, in some cases they even carry less structure than JPEG images.
While we find a notable increase in false negatives for iris recognition, there is barely any impact on fingerprint recognition.
This shows that fine details and complex patterns (as in iris, fabrics, tattoos)
are more susceptible to AI compression than thicker and clearer lines (as in
fingerprints). We hope that this work contributes towards a better
understanding for the implications of advanced compression algorithms for
biometric data in law enforcement tasks.

\bibliographystyle{IEEEtran}
\bibliography{IEEEabrv, submission}

\end{document}